\documentclass{article}
\usepackage{graphicx} 
\usepackage{amsmath}    
\usepackage{amssymb}    
\usepackage{braket}     
\usepackage{hyperref}
\usepackage{booktabs}
\usepackage{tabularx}

\title{Extending Ontologies: From Dense Embeddings to Hybrid Quantum-Fuzzy Systems}
\author{Angjelin Hila}
\date{December 2024, Updated April 2026}

\begin{document}

\maketitle

\begin{abstract}
    LLMs have revolutionized knowledge representation and retrieval, but lack the explicit modeling that knowledge ontologies possess. This paper surveys the ways that ontologies and knowledge graphs have been integrated with dense embedding algorithms. All hitherto attempts involve a trade-off between probabilistic and crisp inference. This paper proposes a novel frontier for devising knowledge representation systems that can simultaneously accommodate probabilistic and crisp inference in the same representation. To this effect, the paper proposes neuro-quantum-fuzzy systems as knowledge representation systems that accommodate both classical and contextual inference implemented through quantum-neural networks (QNN). 
\end{abstract}

\section{Introduction}
The exponential proliferation of human and machine-generated information in the web and beyond necessitate the generation of knowledge representation standards that can facilitate the interoperability of data across systems and platforms, enhance discovery through search and query, and more recently improve the performance of LLMs by improving Generative AI performance through knowledge graphs. Due to the success of deep learning since the early 2010s, word embedding representations, and the ground-breaking transformer architecture that spawned LLMs, knowledge graphs (KGs) have been successfully integrated with dense vector representations. However, the variety of integrations proposed leave a major problem unsolved: transformer-enhanced KGs tend to lose logical structure, undercutting the goal of the full integration of context sensitivity and logical inference. 

The purpose of this paper is to cross-examine two sources of intelligence: the connectionist paradigm of bottom-up learning implemented through deep neural networks and the top-down models of knowledge representation designed by humans that explicitly support logical inference on domain-specific knowledge bases. On the one hand, Generative AI and LLMs suffer from epistemic problems of low trust (though this is increasingly less the case) and hallucinatory behavior stemming from the black box problem \cite{alansari2026large, ribeiro2016why}. On the other hand, Good Old Fashioned AI (GOFAI) suffers from rigid structures, lack of universal standards of representation, and inability to dynamically adapt to new information by learning new rules and features \cite{bengio2013representation}. In this paper, we combine the strengths of these two paradigms toward the problem space of knowledge representation and prediction through flexible knowledge graphs. A more precise way of framing the problem is that our concern here is the construction of knowledge representation systems that balance probabilistic inference with crisp inference, while enabling scaling and interoperability. 

Ontologies are notoriously expensive and suffer from problems of interoperability, ambiguity, inconsistency and incompleteness \cite{fahad2008ontological, dermeval2014systematic}. Recent advances in artificial intelligence (AI) and in particular Natural Language Processing (NLP) offer a blueprint for a family of solutions that leverage artificial neural network (ANNs) learning capacities to improve and extend the functionality of ontologies by expanding actionable capacities. At the same time, ANNs and specialized architectures are beset by lack of access to ground truth and inability to achieve crisp inference that limit their learning capacities. The integration of AI and hand-crafted KOs enables the mutual augmentation of these information systems by cross-pollinating solutions between human-constructed knowledge bases with access to ground truth and computationally powerful deep learning algorithms that extend those knowledge bases by rendering them adaptive and pliable. 

In this paper I explore machine learning (ML) and artificial intelligence (AI) solutions as a means of augmenting the expressivity, consistency, dynamicity, and computational tractability of ontologies. In particular, I explore the application of recent innovations in word embeddings as a means of enriching semantic relationships in ontologies, enabling learning through AI, and knowledge ontology (KO) completion and extension. Word Embedding solutions to Resource Description Framework (RDF) and Web Ontology Language (OWL) have been suggested to varying degrees of success. I extend these solutions by proposing Transformer-power and BERT solutions to Generative AI . I consider, conversely, the application of ontologies to LLMs as a means of improving the performance of the former. The central thesis that I defend is that the integration of KO (Knowledge Ontologies) and LLM capacities can be mutually beneficial. Finally, I consider some novel ideas that include extending the logical syntax of OWL with fuzzy logic, quantum computation, and a process ontology that I foresee defining the future of knowledge representation as an extension of collective rationality.

\section{From Ontology to Ontologies}

While the term ontology has a long history in the field of philosophy concerning topics ranging from existence, being and logic \cite{Ferraris2005Ontology, aristotle2009complete2, VanInwagen2020Metaphysics}, its transposal into the fields of computer and information science is a more recent phenomenon. Historically, the word metaphysics preceded ontology as the name of a collection of fourteen books by Aristotle about “first philosophy” or axioms \cite{aristotle2009complete, VanInwagen2020Metaphysics}. In the early modern era, the meaning of metaphysics expanded into subjects previously understood to be about physics proper. The term ontology was consequently invented to distinguish the expanded usage of metaphysics from its original sense of confinement to being as such \cite{Ferraris2005Ontology, Hofweber2023LogicOntology}.  Etymologically, ontology comprises the amalgam of ontos meaning being and logia meaning study of, deriving from the Ancient Greek logos \cite{Ferraris2005Ontology, VanInwagen2020Metaphysics}. 

The construction and application of formal ontologies in computer and information science derive a great deal of theoretical scaffolding from nineteenth century developments in logic and ontology. The development of first-order logic, the modern standard in science, linguistics, and philosophy, can be imputed primarily to Gotlob Frege who developed the first predicate calculus comprising propositions that quantify variables over non-logical objects \cite{frege1970begriffsschrift}. First-order logic distinguishes from higher-order logics on the basis of quantification over abstract objects. Modern ontology, on the other hand, begins with Leśnieski who distinguished between protothetic, the logic of propositions, ontology, the logic of names and functors, and mereology, the logic of part-whole relations \cite{rickey2011lesniewski, simons1987parts}. 

At first glance, ontologies may appear similar to classification systems and taxonomies. While both ontologies and classification systems involve class hierarchies, some theorists individuate them on the basis of purpose \cite{madsen2009ontologies}. According to Madsen and Thomsen \cite{madsen2009ontologies}, ontologies enable and simplify knowledge representation about phenomena, whereas classification systems subdivide phenomena into classes as a means of ordering them. Unlike classification systems, ontologies encode complex structures that specify flexible relationships between sets, properties and relations \cite{gruber2009ontology}. 

The role of ontologies within information systems spans the formal construction of computer languages as well as the interoperability of database systems through knowledge representation. With respect to the design of computer languages, ontologies define object literals, class hierarchies in object-oriented programming, semantics and inheritance rules \cite{koide2005owl}. Besides general purpose languages, ontologies also play a fundamental role in the design of Domain Specific Languages (DSL) such as Structured Query Language (SQL), Extensive Markup Language (XML), Hardware Description Language (HDL), and Resource Description Language (RDF). The usage of ontologies therefore informs the design of computer languages, algorithmic structures, the design of artificial intelligence, and the design of knowledge representation graphs (KGs) in the context of the semantic web and linked data. 

In the field of Information Science, applied ontology proper begins with the development of the Knowledge Interchange Format (KIF) and the Web Ontology Language (OWL) \cite{arp2015bfo}. In this review, I’m concerned with the application of Natural Language Processing (NLP) representations known as embeddings to ontologies. 

\section{Knowledge Organization}

Knowledge graphs are graph data structures consisting of vertices and edges also termed nodes and links that aim to capture relationships and structural properties between entities. Graph structures can vary widely from undirected to directed, hierarchical, time-based, and hypergraphs, which allow edges to join more than two vertices \cite{barrasa2023knowledge}. While graphs are studied as abstract mathematical structures, in the context of information systems they play a role in representing knowledge bases and structuring unstructured data and resources through metadata. In the latter domain, graph structures model social networks, ICT networks, brain networks and form the basis for the network architecture of artificial neural networks (ANNs). In the context of knowledge organization, ontologies formalize predefined schemas within a graph structure \cite{allemang2011semantic, barrasa2023knowledge}. 

\section{Ontologies and the Semantic Web}

The Resource Description Framework (RDF) and subsequent iterations like RDF 2 and the Web Ontology Language (OWL) were developed as an ontological framework to structure unstructured data on the web and define specialized knowledge graphs that facilitate inference, interoperability, standardization, and complexity modeling \cite{allemang2011semantic}. RDF and OWL underlie ambitious projects like the semantic web, which aim to provide a standard for structuring and making machine-readable data on the web and linked open data, which aims to interlink structured data in standard, machine-readable formats open to anyone \cite{allemang2011semantic}. 

An RDF knowledge graph \( K \) defines a set of triples as subject-predicate-object \( (s, p, o) \), where each element belongs to the sets of resources \( R \), relations \( P \), and object literals \( L \), such that \( (s, p, o) \in R \times P \times (R \cup L) \). Resources constitute the union of entities \( E \) and concepts \( C \), i.e., \( R = C \cup E \) \cite{alshargi2018concept2vec}. RDF treats every node and link as a resource possessing a unique resource identifier (URI). Some problems with the original RDF schema include a combinatorial explosion from representing all data as triples, and an inability to represent complex data structures—such as those requiring indefinite arity or predicate relations, where arity defines the number of arguments in a logical statement. Nevertheless, the RDF schema is extensible and has been used to define richer ontologies such as RDF 2 and the Web Ontology Language (OWL).

OWL extends the logical expressivity of RDF \cite{owl22012overview} by enabling class and property hierarchies, equivalence and disjointness, quantification, cardinality restrictions, the construction of complex classes through operations of union, intersection and complementarity and property characteristics such as transitivity, symmetry and inverseness. Class hierarchies such as subClassOf or subPropertyOf enable inheritance and entailment relations between classes. Class disjointness such as disjointWith enables class mutual exclusivity, whereas class equivalence such as equivalentClass enables classes to share their instances. Cardinality restrictions expand OWL expressivity to represent real world relationships that have cardinality restrictions such as 1-N or N-N relationships. Quantification enables OWL relations to express existential quantification equivalent to at least one, and universal quantification, which defines the predication of a property to all class instances. These representational capacities make OWL a powerful syntactic tool for constructing knowledge ontologies, examples of which include Gene Ontology (GO), Biological Pathway Exchange (BioPAX), and Human Phenotype Ontology (HPO), which successfully embed accumulating biological knowledge within structured ontologies enabling querying, inference, statistical analysis and ML learning on graphs \cite{barrasa2023knowledge}. Examples in other knowledge domains include DBpedia, which maps Wikimedia data into RDF linked-data graphs through crowdsourcing, Wordnet, an English language lexical ontology geared primarily toward synonymy relations, and Friend-of-a-Friend (FOAF), a voluntary-based machine-readable friendship network based on RDF/OWL schemas. 

Applications of machine learning (ML) algorithms to graph data structures include graph-native ML algorithms that extract rules or features from the graph, learn graph dependencies and metadata in order to complete missing data, and finally can be used for prediction and classification tasks. While traditional algorithms like linear and logistic regression, random forest and support vector machines (SVM) can be used, more profitable applications include utilizing artificial neural networks. In particular, I explore the benefits of applying word embedding algorithms and graph neural networks (GNNs) in order to generate a rich vectorized semantics that exploits network topology for prediction, classification and network completion tasks.

\section{Token Embeddings}

The recent success of deep learning enhanced NLP evinced by the emergence of LLMs and Generative AI more broadly stem from two crucial architectural innovations in the field of artificial neural networks (ANNs): contextual word embeddings and the Transformer architecture. Contextual Word Embeddings mark a departure from the bag of words (BoW) model where words are encoded as sparse vectors and fed into the model without regard for word order \cite{zhang2010bow}. 

In order to perform machine learning on text data, linguistic units must be converted into machine-readable representations. The simplest method involves assigning a unique integer to each word in the vocabulary termed integer encoding \cite{grimmer2022text}. However, since machines can mistake the integer assignment for ordinal data, a better method involves representing each word through a sparse vector whose length is equal to the vocabulary. One-hot encoding constructs a square matrix where each word is encoded as a vector of zeros save one unit value whose position in the vector uniquely determines that word within the total corpus. One-hot encoding is particularly useful in classification tasks where the number of classes is typically small. However, it proves to be a computationally expensive method of vocabulary encoding since the vectors increase in length proportional to the vocabulary \cite{grimmer2022text}. These word representation methods are often used in bag-of-words  (BoW) NLP models. BoW models treat the text as an unordered set. The obvious drawback is that natural languages encode meaning through ordinal sequences where word order determines the syntactic function as well as semantic meaning of words. 

The development of recurrent neural networks (RNN) in the 1990s attempted to ameliorate this problem by using feedback to remember ordered sequences \cite{schuster1997brnn}. RNNs and subsequent variants such as Long Short-Term Memory (LSTM) networks employ feedback and a type of memory called a hidden state in order to store the previous output, which is subsequently fed as input in the next iteration \cite{graves2012lstm}. The use of feedback enables RNNs to remember ordered sequences of data, which allows them to learn semantic and syntactic dependencies between words. In a basic recurrent network, the second output is generated by the input while the previous output is stored as an n-dimensional vector called the hidden state \cite{medsker1999rnn, grossberg2013rnn}. 

While RNNs and more sophisticated variants like LSTMs that utilize an additional vector called the cell state to expand model memory \cite{hochreiter1997lstm, vanhoudt2020lstmreview, graves2012lstm} represent an improvement upon BoW, they still process textual information sequentially and rely on encoding dependencies within a single vector, which limits the context window to broadly local dependencies and hinders the network from learning longer-term dependencies that describe global properties embedded within a corpus of text \cite{graves2012lstm}. Therefore, while RNNs improved upon the unordered BoW model by processing texts sequentially, the traditional encoding methods they employed did not capture word relationships. 

Aiming to transcend these limitations, word embeddings constitute an alternative method of word encodings that utilize distributed representations to represent words as dense vectors \cite{rumelhart1986pdp, mikolov2013distributed, mikolov2013efficient}. Word embeddings constitute a departure from the traditional bag-of-words word representations wherein words are encoded as unique integers or sparse, one-hot, n-hot, frequency or TF-IDF encoded vectors that do not map word interrelationships. 

Embeddings represent a vocabulary as an n-dimensional space by assigning each word a relative position through a distance measure between words typically computed through cosine similarity. Cosine similarity captures the relative correlations between words as the angle between two word vectors such that the smaller the angle, the higher the similarity. Depending on the domain of discourse, the context window size that defines the nearness threshold as well as the dimensionality (where dimensionality denotes vector cardinality) of the word embedding can vary, though low dimensional embeddings have proven to be highly effective \cite{grimmer2022text}. 

\[
\cos(\theta) = \frac{\mathbf{A} \cdot \mathbf{B}}{\|\mathbf{A}\| \, \|\mathbf{B}\|} = \frac{\sum_{i=1}^{n} A_i B_i}{\sqrt{\sum_{i=1}^{n} A_i^2} \, \sqrt{\sum_{i=1}^{n} B_i^2}}
\]

Mikolov’s Word2Vec algorithm comprises two opposite models of learning word context: the Continuous Bag of Words Model (CBOW) and the Skip-Gram models \cite{mikolov2013efficient}. CBOW predicts a target word based on the surrounding context. Skip-Gram predicts the context based on the target word \cite{mikolov2013efficient}. Both models learn word contexts iteratively instead of taking the entire corpus into account with one shot \cite{mikolov2013efficient}.

While embeddings capture word similarities, they do not discriminate between contextual meanings that arise from polysemy, for example. This problem has been addressed by Embeddings from Language Models (ELMo) which encode not only the meaning of the word in the wider corpus but also its meaning relative to the words most proximate to it within particular instantiations \cite{babcock2021generative}. Contextual word embeddings capture associative strengths between words as surrogates for lexical meaning as well as connotative meanings across use-contexts. 

Global Vectors for Global Word Representation (GloVe) proposed by Pennington et al. in 2014 aims to improve upon Word2Vec by incorporating Latent Semantic Analysis (LSA) in order to capture global contexts \cite{pennington2014glove}. GloVe computes the co-occurrence matrix of each word in the corpus within a fixed and often symmetrical context window and uses weighted least-squares regression with stochastic gradient descent to train low-dimensional vectors that capture each word's global meaning relative to the entire corpus \cite{pennington2014glove}. While both Word2Vec and GloVe use words as n-gram tokens, FastText, proposed by Bojanowski et al. \cite{bojanowski2017subword}, decomposes words into a sub-word n-grams and therefore learns vector representations at a higher level of granularity. FastText has been demonstrated to work better with out-of-vocabulary words; namely, words not found in the vocabulary embedding. 

Unlike Word2Vec and GloVe, Embeddings from Language Models (ELMo) generates deep contextual embeddings using bidirectional LSTM that capture different embeddings for different words across a variety of contexts \cite{peters2018deep}. Specifically, ELMo leverage a convolutional neural network (CNN) to learn word vectors from character-level tokens, which it subsequently passes through a forward and backward LSTM. Because ELMo use sentences as contextual boundaries, the model learns unique word vectors per sentence, producing a large output vocabulary and forfeiting the stability of lexical meaning \cite{peters2018deep}. While the original architecture leverages a hierarchy of embeddings, where the lower CNN embeddings function as stable lexical units, and higher LSTM embeddings function as contextual meanings, ELMos are prohibitively computationally expensive and were, as a result, quickly eclipsed by the emergence and dominance of transformer-based models \cite{babcock2021generative}. 

This brings us closer to the state-of-the-art, namely the emergence of transformer models, which rendered the afore-explicated models, more or less, obsolete. Transformer models solve the problems of both static and sequential models such as RNNs and LSTMs by dynamically parallelizing the vector learning process \cite{vaswani2017attention}. Dispensing with recurrence and convolutional components, the transformer computes the importance of every word to every other word in an input sequence without requiring a hidden state and not treating strings as ordinal sequences during the learning process \cite{vaswani2017attention}. 

The transformer achieves this through three parallel matrices, typically framed as the query ($Q$), key ($K$), and value ($V$) matrices. Each matrix is the product of the input matrix $X$ and initialized weights $w$. To learn the relevance of a word to another in a sentence, the transformer multiplies the $Q$ and $K$ matrices together, which yields a square matrix with the dot-products of every word to each other, from which pairwise cosine similarities can be obtained \cite{vaswani2017attention}. This product matrix is divided by the square root of the dimensions of the $K$ matrix (which typically is a fraction of the total dimensionality) and normalized through softmax into probabilities, which are subsequently converted into the final attention embedding by multiplying the normalized matrix to the value matrix $V$ \cite{vaswani2017attention}. Because multiple attention heads are used, the transformer also involves a reassembly step in which the independent attention heads are concatenated together and reintegrated into a matrix of the full dimensionality size \cite{vaswani2017attention}. 

Finally, the output matrix is fed through a feed-forward network with an activation function that attends to each token independently and applies backpropagation to readjust the weights of each of our $Q$, $K$, and $V$ matrices. To reconstruct an ordinal sequence, the transformer uses positional encodings, which assign each token a position in a sequence \cite{vaswani2017attention}. The innovative achievement of the transformer architecture is that the number of rows in each attention head constitutes the context size that dictates how many tokens attend to each other simultaneously \cite{vaswani2017attention}. Since the introduction of transformer architecture, context windows have ballooned from 512 to more than 2 million tokens \cite{kilpatrick2024gemini}. The formula for a transformer attention head is provided below: 

\[
\text{Attention}(Q, K, V) = \text{softmax} \left( \frac{QK^T}{\sqrt{d_k}} \right) V
\]

\section{Ontologies and Embeddings}

The application of embeddings to KGs to improve the performance of ontologies has been proposed. Ristoski and Paulheim \cite{ristoski2016rdf2vec} first proposed RDF2vec, the application of embeddings to RDF schemas. They find that applying Mikolov et al’s Word2Vec model \cite{mikolov2013distributed} yields vector representations of RDF graphs that outperform other techniques for the propositionalization of RDF graphs for machine learning tasks. In particular, they use two techniques to convert RDF graphs from DBpedia and Wikidata into sequences: Weisfeiler-Lehman Subtree RDF Graph Kernels (K2V) and graph walks (W2V). They additionally apply both the CBOW and SkipGram embedding models on the data and show that K2V with SkipGram outperforms the baseline graph feature-expression models on traditional regression and classification tasks \cite{ristoski2016rdf2vec}. 

Node2Vec introduced biased walks such that walks can bias local connections or global properties \cite{grover2016node2vec}. Node2Vec learns continuous feature representations of nodes through mapping of nodes to a low-dimensional space of features that maximizes the likelihood of preserving node network neighborhoods \cite{grover2016node2vec}. Node2Vec introduces biased random walks on general graphs that include a return parameter that encourages the walk to return to the previous node for local exploration and an in-out parameter that encourages global exploration of the graph. These parameters flexibly balance homophily (tightly clustered nodes) and structural equivalence between unrelated nodes. 

Building on these approaches, Smaili et al \cite{smaili2019opa2vec} first proposed the application of embeddings to knowledge ontologies in biomedical science. They proposed OPA2Vec to generate vector representations of biological entities in ontologies by applying a Word2Vec model pre-trained on either a corpus of abstracts or full-text articles to produce feature vectors from collected data. They used their trained model in protein-protein interaction and gene-disease association prediction tasks based on phenotypic ontology and found that OPA2Vec outperformed existing models. In OPA2Vec the set of axioms in the ontology are represented as a document of sentences modeled by Word2Vec \cite{holter2019owl2vec}. 

Following OPA2Vec and Onto2Vec \cite{smaili2018onto2vec}, Holter et al \cite{holter2019owl2vec} first proposed OWL2Vec in order to expand the application of embeddings from instances to lexical information. Holter et al \cite{holter2019owl2vec} identify limitations with these approaches like the generation of noise by OWL constructs, inability to discriminate between relationships like SubClassOf and DisjointWith as well as embedding limitations for small-to-medium sizes ontologies with modest corpuses. Inspired by Node2Vec, they propose to solve these problems through a four step approach that generates semantic embeddings from the ontology : (i) project the ontology into a graph, (ii) implement several strategies to walk the ontology graph, (iii) creates a corpus of sentences according to the walking strategies, and (iv) generate concept embeddings from that corpus \cite{holter2019owl2vec}. They show that flexible graph walking strategies to walk biasing through edge weighting produces superior semantic relationships that scale better than RDF2Vec and Onto2V \cite{holter2019owl2vec}.

More recently, Chen et al have \cite{chen2021owl2vec} extended OWL2Vec called OWL2Vec* that incorporates lexical information into the embedding process such as labels, comments, RDFs, in order to capture natural language descriptions in addition to structural relationships. Unlike OWL2Vec, OWL2Vec* explicitly encodes logical constructors such as class hierarchies, quantification and class equivalence and disjointness that endow OWL with high expressivity as an ontology. OWL2Vec* produces a hybridized structure and semantic embedding document that correlates semantics with graph structures with superior performance on class-membership and class subsumption prediction. Evaluating OWL2Vec* in the tasks of class-membership and class subsumption prediction on ontologies like Gene Ontology (GO), and Food Ontology (FoodON), Chen et al find that it outperforms the aforementioned models as well as Transformer-based classifiers \cite{chen2021owl2vec}.  

A rival method of learning latent representations from graphs include applying Graph Neural Networks (GNNs) to ontologies \cite{mezar2022ontology, hohenecker2020ontology}. GNNs learn node and relationship representations dynamically by propagating through a graph. They utilize end-to-end learning that involves a direct mapping from input to output trained at specific tasks like link prediction and node classification. For this reason, they excel at structural analysis and graph completion as demonstrated by Mežnar et al \cite{mezar2022ontology}. However, they rely heavily on graph structure rather than explicit semantics and therefore have trouble differentiating the semantic meaning of ontology predicates. 

As shown, neuro-embeddings on ontologies offer an array of advantages including latent semantic understanding, contextual variability and ontology completion, reusability in ML tasks, improved information retrieval, and progress toward ontology alignment projects that seek to map concepts between ontologies as well as integrate heterogeneous data structures. Despite these advantages, embeddings trade-off the symbolic robustness of ontologies and KGs resulting in loss of reasoning capabilities reliant on its explicit semantics and attendant logical expressivity. Furthermore, embeddings abstract away the dynamic flexibility of ontologies by projecting them onto static/rigid vector representations. Below I delve into the SOTA advances in neuro-KGs since the advent of transformers, and in the subsequent section I propose novel solutions to these losses that involve leveraging fuzzy reasoning and quantum logic. 

\section{Attention-based KGs}
Since the advent of the transformer architecture, hybrid models that aim to integrate transformer embeddings with KGs have emerged \cite{Chen2021HittER, Liu2022MaskAndReason, Yao2019KGBERT, Wang2022SimKGC} that build upon earlier embedding-enhanced KGs \cite{Wang2014TransH, Ji2015TransD, Lin2015TransR}. Here, we highlight three methods that represent distinct ways that KGs, transformers and LLMs have been combined to solve specific problems. This overview will circumvent feature-level integration, where embeddings are adjoined as features to an existing graph, a method that I have utilized in my work. Because feature-level integration of transformer encodings conserve graph topology, these methods are better understood as structural-semantic graphs.

\subsection{KG-Aware Transformers}
KG-aware transformers refer to transformers that are trained on knowledge graph structure such as RDF triples \cite{Liu2024KnowFormer}. Liu et al \cite{Liu2024KnowFormer} proposed Knowformer, a transformer-based architecture that operates on KG triplets by modifying each $QKV$ matrix into KG-structurally-aware embeddings that can accurately predict KG completion queries consisting of head-entities $h$, tail-entities $t$, and relations $r$: $(h,r,t)$ \cite{Liu2024KnowFormer}. Knowformer first converts discrete KG triplets into initialized continuous vector representations that encode relation-conditioned vector contributions between nodes \cite{Liu2024KnowFormer}. Given a directed triple Obama--Nationality--USA, Knowformer produces a relation-conditioned update to USA's embedding from the Obama--nationality entity-relation pair. In aggregate, the USA embedding absorbs conditioning from observed triples that contain USA as the tail entity. The corpus-wide KG-conditioned entity embeddings are subsequently routed through a modified attention mechanism that uses a positive kernel that substitutes the full pairwise softmax function. The chief innovation of Knowformer, therefore, is to infuse the input matrices in the kernalized transformer with graph-structured inductive bias. The reconstructed attention mechanism below replaces the exponential softmax normalization with a similarity kernel that positively shifts the dot product between the representations of entity $u$ and $v$ by adding $1$ to it, normalizing it by the total sample space, which then weights the value of the representation of entity $v$:

\[
\operatorname{Attn}(x_u)
=
\sum_{v \in \mathcal{V}}
\frac{
\kappa\left(f_q(x_u), f_q(x_v)\right)
}{
\sum_{w \in \mathcal{V}}
\kappa\left(f_q(x_u), f_q(x_w)\right)
}
f_v(x_v).
\]

Knowformer learns KG structural representations by encoding the input matrices with prior graph structure observed from KG distributed relations. The modified transformer mechanism learns global contextualized vectors from the structure Knwoformer inherits from its observed graph triplets, yielding a representation that excels at KG completion and link prediction. However, Knowformer nevertheless forfeits the crisp inferential properties that characterize KGs. KG-aware transformers learn probabilistic representations of typed KG relations that enable prediction, generalization, and flexible extension of KG structure at the expense of logical properties such as class hierarchy, disjointness, quantification and entailment.

\subsection{Relational Graph Transformers}
Instead of learning KG structure through the transformer model, Relphermer, short for relational graph transformers, modifies the attention logits themselves by interpolating relation types, structure and neighborhood encodings \cite{Bi2024Relphormer}. This enables relational graph transformers to map more explicit features of KGs into a transformer-encoded sequence. Proposed by Bi et al \cite{Bi2024Relphormer}, Relphormer leverages a three-step process to inject structural properties into the attention mechanism toward KG completion, Question Answering, and recommendation tasks: (a) contextual subgraph sampling method called Triple2Seq that converts input tokens (treated as node types: entity nodes, relation nodes etc) into sequential representations, (b) a modified attention mechanism that adds a structural bias to the QK matrix, and (c) a masked knowledge modeling task to measure contextual loss \cite{Bi2024Relphormer}. 

\[
a_{ij}
=
\frac{
(\mathbf{h}_i W_Q)
(\mathbf{h}_j W_K)^{\top}
}{
\sqrt{d_k}
}
+
\phi(i,j).
\]

The bias term $\phi(i,j)$ encodes a structure vector of the normalized adjacency matrix of the sample subgraph, where each power denotes walks $>$ length 2: 

\[
\mathbf{s}_{ij}
=
\left[
\widetilde{A}_{ij},
(\widetilde{A}^{2})_{ij},
\ldots,
(\widetilde{A}^{m})_{ij}
\right],
\qquad
\phi(i,j)
=
f_{\mathrm{str}}(\mathbf{s}_{ij}).
\]

In conjunction with masked language modeling (MLM), the structural bias provided by the subgraph adjacency matrix enables Relphormer to learn node type identity, namely whether a token is an entity or a relation. The subgraph bias conditions the probability distribution over candidate entities in the MLM task. The attention bias mechanism enables Relphormer to learn node identity and local graph context, but the model ultimately forfeits class disjointness, quantification, entailment, cardinality, and formal class semantics. 

\subsection{KG-Enhanced LLMs}
Another line of integration stems from the aspiration of enabling transformer representations to update their knowledge bases based on editing of facts. For example, proposed by Zhang et al \cite{Zhang2024KnowledgeGLAME}, GLAME constitutes a KG-aided LLM editing method with the goal of enabling LLMs to update downstream effects on their internal representations. GLAME involves two steps: (a) knowledge graph augmentation (KGA), which constructs a subgraph around the LLM edit, and (b) graph-based knowledge edit (GKE), which injects edited and associated knowledge into LLM parameters \cite{Zhang2024KnowledgeGLAME}. GLAME uses a locate-then-edit technique that extracts hidden state vectors standing for an entity/subject from an LLM in its feed-forward network (FFN) stage to train an edited KG subgraph via relational graph convolutional network (RCGN) \cite{Zhang2024KnowledgeGLAME}. The KG subgraph encodes downstream inference, the hidden subject key located through LLM prompting represents the target entity, and the RCGN training encodes a new value that is subsequently reinserted into the FFN \cite{Zhang2024KnowledgeGLAME}. GLAME is then evaluated using factual editing benchmarks such as 
\texttt{COUNTERFACT}, \texttt{COUNTERFACTPLUS}, and \texttt{MQUAKE} \cite{Zhang2024KnowledgeGLAME}. GLAME's relationally-aware model editing and inferential propagation shows that latent LLM representations can be precision targeted for knowledge revision. However, while GLAME shows remarkable precision in editing relevant inferential effects while leaving unrelated facts constant or ceteris paribus in the surrounding neighborhood, it is unclear whether this method produces wider untested effects on the global network. Further, it opens up questions about the potential for editing abstract representations rather than merely facts through KG structure.  

These attempts are heavily redolent of Quine's meaning holism as espoused in his seminal \textit{Two Dogmas of Empiricism} \cite{Quine1951TwoDogmas}. Quine argued that revisions to beliefs that are contingent on empirical facts occasion revisions of beliefs that are contingent on other beliefs. He envisioned an inferential web with centrifugal and centripetal extremities: beliefs that impinge upon the tribunal of experience and beliefs closer to the center that are inferentially dependent on surrounding beliefs \cite{Quine1951TwoDogmas}. Quine's proposal had the radical consequence that all beliefs, even logical truth, maintained a relation, albeit indirect, to empirical input \cite{Quine1951TwoDogmas}. A computational representation of Quine's web would require an integration of semantic representations a la distributional semantics as captured dynamically by the transformer architecture, and logical relationships that enable inferential updates that adhere to necessity and sufficiency criteria. 

\subsection{Agentic KG-Enhanced LLMs}
Another line of integration constitutes agentic approaches that combine KG structure and dense embeddings. GraphRAG \cite{Edge2024GraphRAG} represents a means of improving traditional retrieval augmented generation (RAG) by constructing graph structures that improve retrieval, reasoning and in a more limited way factual verification. GrapRAG proceeds in two stages: (a) an indexing stage that builds a KG from a pre-selected corpus and (b) a querying stage that retrieves relevant data embedded in the graph structure \cite{Edge2024GraphRAG}. The indexing stage partitions documents into chunks tracked through metadata and extracts entities, relationships and primitive assertion objects from which it builds a graph structure that enables statistical analysis across representational units: nodes, edges, communities, and at the scale of the global graph \cite{Edge2024GraphRAG}. GraphRAG deploys an LLM to extract graph components and derive summaries at various levels of analysis, which are used during the query stage to provide answers that leverage graph structure instead of merely a contextual gradient \cite{Edge2024GraphRAG}. Relying on LLMs to extract graph components underscores an epistemic weakness of GraphRAG in addition to relying on a preselected document corpus, whose validity needs to be vetted independently. 

In aggregate, these methods achieve important KG integrations with transformer representations and LLMs but broadly forfeit key graph structures that confer on KGs robust knowledge transmission. Table \ref{tab:kg-transformer-property-comparison} summarizes the forfeited-KG properties of the above described innovations. A proper integration of discrete representations that support crisp inference with contextual and inherently probabilistic inference requires an architecture that hierarchically scaffolds from associationist semantics to higher-order semantics that take those lower-order representations as inputs. Second-order representations could represent hierarchical dependencies that enable downstream inferences on world-knowledge. 

\begin{table}[t]
\centering
\small
\renewcommand{\arraystretch}{1.18}
\begin{tabularx}{\textwidth}{p{0.20\textwidth}XXX}
\toprule
\textbf{Primitive KG property} & \textbf{KnowFormer} & \textbf{Relphormer} & \textbf{GLAME} \\
\midrule
Entity identity & Yes & Yes & Yes \\
Relation identity & Yes & Yes & Yes \\
Directionality & Yes & Mostly yes & Partially yes \\
Arity & Mostly binary triples & Mostly triple/subgraph-based & Mostly binary KG triples \\
Class membership & Only if encoded as relation & Only if encoded as relation & Only if retrieved from KG \\
Class hierarchy & Learnable pattern, not enforced & Learnable pattern, not enforced & Usable if in subgraph, not enforced \\
Domain/range & Not enforced & Not enforced & Not enforced \\
Disjointness & Not enforced & Not enforced & Not enforced \\
Quantification & No & No & No \\
Cardinality & No & No & No \\
Entailment & Plausibility ranking & Masked prediction/link prediction & Multi-hop QA portability \\
Negation & Not robust/formal & Not robust/formal & Not robust/formal \\
Provenance & No, beyond dataset split & No, beyond dataset split & Weak; external KG source, not full provenance \\
Temporality & No, unless encoded as triples & No, unless encoded as triples & No, unless KG/query encodes it \\
Modality/status & No & No & No \\
\bottomrule
\end{tabularx}
\caption{Comparison of primitive knowledge-graph properties preserved, approximated, or weakened in KnowFormer, Relphormer, and GLAME. The table distinguishes learned relational regularities from formally enforced knowledge-representation semantics.}
\label{tab:kg-transformer-property-comparison}
\end{table}

\section{Beyond Embeddings: \\
Deep Neuro-Fuzzy Quantum Systems}

My proposal for increasing the representational power, modularity, and inference capabilities of ontologies in order to enhance the semantic web and KGs as well as foster interoperability across systems involves hybridizing quantum and fuzzy logic within embedded ontologies. 

Quantum and fuzzy logic diverge along domains of applicability and evolved to solve different problems but share important homologies that may be dually exploited for ontological representation. Quantum logic exploits properties of quantum states like superposition and entanglement in order to perform classically intractable computations. Challenges with quantum logic reside less with formal design than engineering constraints necessary for the conservation of quantum superpositions. Near-zero kelvin temperatures are required to perform computations that maintain the state of superposition where all possible states of the system obtain simultaneously as described by Schrodinger’s wave function \cite{cleri2024quantum}. Quantum logic leverages quantum states to perform logical operations that exceed the polynomial time O(N), O(N2)... O(N3) limits of classical computers. For example, cryptographic deciphering requires exponential time algorithms expressed in O notation as O(2n). 

Fuzzy logic was developed to overcome the limitation of classical logic to the principle of bivalence where truth functions yield binary values. Fuzzy logic introduces multivalued outputs which are typically normalized as real-numbered values within the unit interval [0,1]. The choice of membership function varies across use-cases, though a Gaussian function that produces a differentiable symmetrical curve is suited as a representation of gradience \cite{klir2015fuzzy}. Multivalued logics accommodate uncertainty and degrees of truth, which have applicability in engineering, control theory, and machine learning \cite{klir2015fuzzy}. Fuzzy logic entails fuzzy set theoretic operations that formalize class membership as a gradient. This means that class membership is not a matter of either-or like in classical logic, but a matter of gradation: some instances belong more to a class than others. As such, fuzzy logic subsumes classical logic as a special case. In contrast to classical logic, fuzzy logic better models semantic flexibility and class membership gradience exhibited in natural languages. 

Several crucial differences distinguish quantum from fuzzy logic. While in a state of superposition, a qubit can occupy a continuum—or infinite number—of combinations of its basis states \( \ket{0} \) and \( \ket{1} \) (representing the unit vectors \([1\ 0]\) and \([0\ 1]\)), upon observation the system always collapses into a discrete outcome of either basis vector. This contrasts with fuzzy logic, where truth values can continuously range over the unit interval \([0, 1]\).

Furthermore, fuzzy logic assumes the axiom of commutativity, such that
\[
A \cup B = B \cup A,
\]
whereas quantum logic violates commutativity:
\[
A \cup B \neq B \cup A,
\]
since logical operations are represented via matrix multiplication, which is non-commutative (i.e., the number of columns in the left matrix must equal the number of rows in the right).

Quantum logic also violates the distributive law:
\[
A \cup (B \cup C) \neq (A \cup B) \cap (B \cup C).
\]

While the amalgamation of fuzzy systems with connectionist architectures was proposed as early as 1993 by Jang, who developed a model called adaptive-network-based fuzzy inference system (ANFIS) \cite{jang1993anfis}, the hybridization of fuzzy systems and quantum logic constitutes, to the author’s knowledge, a foray into novel territory. While this approach presents prospects for the next generation of knowledge representation models, the limitations of this approach are also considered given the aforementioned formal differences between fuzzy and quantum logic. It is further worth noting that the application of fuzzy semantics to RDF graphs independent of deep neural networks (DNNs) was first developed by Lv et al. \cite{lv2008fuzzyrdf}. While this and other extensions of fuzzy systems into OWL and KG construction constitute a step in the right direction, we believe that the integration of ontologies and KGs with fuzzy inference and quantum logical operations solve several outstanding problems with present ontologies including computational intractability, complex inference, and scalability. 

According to Cross and Chen \cite{cross2018fuzzy}, fuzziness can be defined a priori within the concept hierarchy or implemented a posteriori to a crisp hierarchy by considering relevant data sources where an instance in a set can partake simultaneously in crisp as well as fuzzy set extensions. Therefore, because fuzzy concepts can be fuzzy subconcepts of supersets, generalization rules are required where an instance $i$ in $N$ with membership degree $u_N(i)$ requires $u_N(i) \le u_M(i)$ (where $M$ denotes the superconcept) so that the membership degree within the subconcept does not exceed the membership degree within the superconcept \cite{cross2018fuzzy}. Fuzzy ontologies like FuzzyDL, Fuzzy OWL 2 therefore require intricate formal rule setting in order to preserve the logical integrity of the ontology making its structure fundamentally static, unscalable, and lacking the adaptability and automated learning inherent within ANNs.

Recent models explore these possibilities within a family of Deep Neuro-Fuzzy Systems (DNFS), where Jang’s original model has gained renewed attention. Talpur et al. (2023) distinguish between three hybrid architectures that combine fuzzy systems and DNNs: sequential, parallel, and cooperative, finding that sequential architectures dominate, while noting that the latter two yield faster and superior performance. DNFS systems instantiate interpretability-accuracy trade-offs, where high accuracy correlates with difficulty of interpreting outputs. This is a problem where the logical inference steps need to be explicitly understood and mapped. Mathematically, they can cause rule explosion from large inputs as a result of the combinatorics of fuzzy inference \cite{mendel1999comments}. Further, DNFSs may be difficult to tune since the tuning methods differ substantially between fuzzy logic and ANNs with the former requiring precise rules and the latter hyperparameter optimization. 

In the sphere of quantum representation, Garg et al. \cite{garg2019quantum} have proposed quantum logic-inspired embeddings to knowledge graphs called Embed2Reason (E2R) aimed to enhance reasoning and inference processes. This method aims to improve the loss of representational crispness that traditional embeddings incur on KG in favor of gaining latent semantics and contextual richness. E2R embeds a knowledge base (KB) into a finite-dimensional vector space where the logical structure is preserved as a lattice of subspaces \cite{garg2019quantum}. Entities are represented as entities in a vector space and membership is determined by whether they fall within the vector’s subspace. The axioms of quantum logic guide the embedding process, ensuring that logical operations can be performed directly over vectors. E2R leverages non-distributivity to preserve class hierarchies in KGs, which embeddings lose in favor of mapping statistical semantic relationships \cite{garg2019quantum}. 

In light of these developments, I propose the integration of quantum computation with Deep Neuro-Fuzzy Systems in order to solve two outstanding problems within ontologies and KGs: computational complexity and equilibrizing robust inference with semantic richness. To the author’s knowledge such a solution has not been proposed toward knowledge ontologies. However, academic attempts to integrate fuzzy and quantum logic exist \cite{mannucci2006blending, gentili2021fuzzy, pourabdollah2022fuzzy}, though they are in their infancy and far from established. The primary thrust of this proposal involves quantum operations on fuzzy-sets through quantum search algorithms. In the context of deep neuro-fuzzy systems this means leveraging quantum computational parallel processing power to solve complex classification problems that involve n-dimensional graded membership spaces. Schmitt et al \cite{schmitt2009fuzzyquantum} suggest a way forward that algebraic products and sums in fuzzy logic align with quantum logical operations under commuting projectors. Because fuzzy logic lacks a mechanism for testing disjointness and overlap, quantum logic can be leveraged to test condition disjointness in order to ensure compatibility with Boolean operators \cite{schmitt2009fuzzyquantum}. Quantum systems are particularly suited to solve the computational complexity that results from deep neuro-fuzzy systems beset by the curse of dimensionality. Standardization of a deep neuro-fuzzy framework remains a problem for realizing these ambitions, but this may be achieved through further work on the latter. 

A blueprint for combining fuzzy operations with quantum computational operations proceeds as follows: states of superposition can be used to represent fuzzy membership by mapping fuzzy sets onto superimposed qubits. Recalling that we represent a qubit using two orthonormal basis vectors, when a qubit is in superposition, its squared probability amplitudes are each \( 0.5 \), represented by the linear combination \( \alpha \ket{0} + \beta \ket{1} \), where \( \alpha \) and \( \beta \) are complex probability amplitudes such that \( |\alpha|^2 + |\beta|^2 = 1 \), and \( \ket{0} \) and \( \ket{1} \) represent the standard basis vectors corresponding to classical bits 0 and 1. Furthermore, multiple qubits are represented via their tensor products.

Input fuzzy sets encoded through basis vectors can be arranged in states of superposition by applying the Hadamard gate, a \( 2 \times 2 \) unitary matrix defined as
\[
H = \frac{1}{\sqrt{2}} \begin{bmatrix} 1 & 1 \\ 1 & -1 \end{bmatrix},
\]
which ensures reversibility due to its orthogonality. When a qubit is in superposition, it can occupy a continuous spectrum of states. However, quantum measurement collapses the qubit into one of the two basis states \( \ket{0} \) or \( \ket{1} \), meaning that membership gradience can only be represented prior to measurement. To operate on such fuzzy superpositions, one can apply reversible logic gates. One method to recover the underlying distribution is by performing repeated measurements to estimate frequency counts as proxies for fuzzy membership values.

To approximate the probabilities associated with superimposed states, we repeat identical experimental setups and use empirical frequencies of measurement outcomes as estimators of fuzzy membership. For example, the quantum state
\[
\psi = \sqrt{0.3} \ket{0} + \sqrt{0.7} \ket{1}
\]
assigns probability 0.3 to the \( \ket{0} \) basis state and 0.7 to \( \ket{1} \). While the state collapses upon a single measurement, repeated measurements allow us to approximate the original distribution by analyzing the resulting frequency histogram.

We implement quantum operations on fuzzy sets through reversible gates such as the controlled-rotation (CROT) gate for fuzzy AND operations, and the phase-flip (Z) gate for fuzzy NOT. Since logical functional completeness guarantees that all logical expressions can be constructed using two operators, it is sufficient to define two reversible quantum gates—or alternatively employ a universal quantum gate such as the TOFFOLI gate. However, fuzzy logic requires operations that account for graded membership, such as fuzzy AND (\( \min(x, y) \)) and fuzzy OR (\( \max(x, y) \)), where \( x \) and \( y \) are membership degrees of distinct fuzzy sets. For instance, if \( x \) denotes the membership degree of the fuzzy set \emph{happy} and \( y \) that of \emph{educated}, then their union can be modeled using \( \min(x, y) \). As the TOFFOLI gate primarily performs conditional flips, additional gates such as CROT, quantum adders, and renormalization gates are needed to implement fuzzy operations and to maintain the normalization condition \( \|\alpha\|^2 + \|\beta\|^2 = 1 \).

In order to complete our model, we adduce a quantum neural layer (QNN) to learn fuzzy membership functions between sets. The QNN utilizes an optimization algorithm like backpropagation to learn the membership functions. Since we require quantum measurement to defuzzify the outputs, we need a refuzzification step before feeding the data to the QNN and a defuzzification step after QNN optimization.  To give an example, the interaction between two proteins can be represented through a fuzzy set where the membership function defines the interaction uncertainty as a normalized real numbered value. Since we’re using quantum computational operations in states of superposition, we extend the number of parameters (e.g. protein interactions) indefinitely in order to compute the interaction matrix without the polynomial time limits imposed by classical computers on computational time-scales. Broadly speaking, the quantum neuro-fuzzy pipeline enables the processing of high-dimensional data by learning otherwise intractable fuzzy membership functions and computing complex inference that conserves logical robustness. An outstanding problem in realizing this pipeline concerns the current limitations of QNNs, particularly the lack of an adequate substitute to the nonlinear activation function that enables ANNs to learn \cite{schuld2014quest}. 

\begin{figure}[t]
  \centering
  \includegraphics[width=\linewidth]{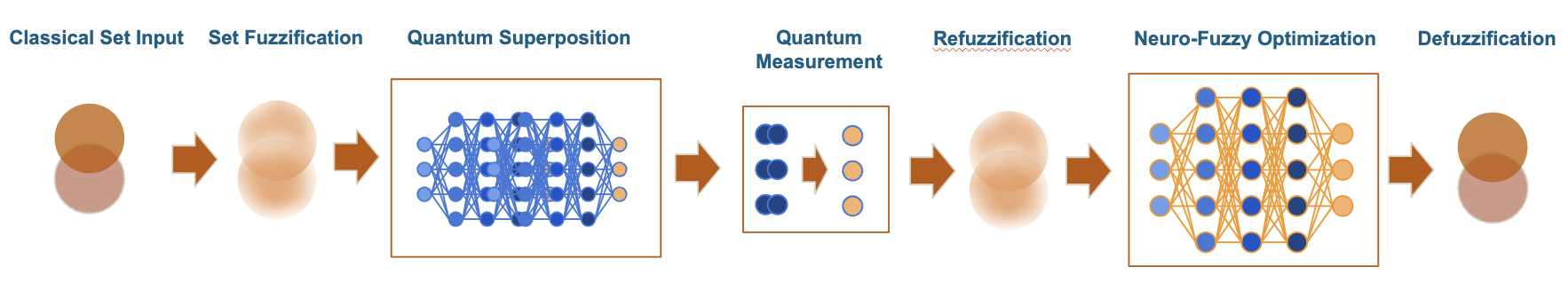}
  \caption{Quantum Neuro-Fuzzy System Pipeline.}
  \label{fig:quantum_neurofuzzy_system}
\end{figure}

A generalization of the proposal to successfully hybridize deep neuro-fuzzy systems and quantum computing entails envisioning dynamic ontologies as a natural evolutionary step in the domain of knowledge representation as a subset of computing systems and Artificial Intelligence (AI). Before deep learning emerged as the dominant paradigm of AI in the 2010s, statistical relational learning (SRL) vied for that position. Models like Bayesian Logic Programming (BLP) and Markov Logic Networks (MLN) demonstrated high performance in tasks that require modeling uncertainty in relational modeling such as causal and noncausal networks \cite{getoor2019srl}. BLP combines Bayesian inference with logic programming, whereas MLN combines Markov Random Fields with first-order logic and weighted formulas to yield learning capability in structured and unstructured domains. These models transcend the limitations of traditional statistical methods that treat data points as identically and independently distributed (i.i.d.) by modeling dependencies between entities \cite{khosravi2010survey}. However, the rise of big data oversaw disproportional performance gains in ANNs and deep learning by outperforming SRLs in scalability, adaptability, and automatic feature extraction. These gains, however, come with obscured explainability, loss of logical ontology, and inordinate data requirements. My emphasis on ontologies resuscitates the utility of explicit symbolic modeling within the connectionist AI paradigm. Despite the accuracy and reliability of ANNs, the internal modeling distributed in the hidden layers is inscrutable to human reasoners and renders their accuracy unexplainable without mentioning tendencies to become unmoored from their evidence-base through hallucinations. It is for this reason, among others, that I have advocated for the synthesis of these paradigms within Neuro-Symbolic AI as the optimum between the poles of top-down and bottom-learning. 

As ontologies and KGs aim to encode complex information about the real world, especially in empirical domains like biology, engineering, and chemistry where complexity reigns, neuro-fuzzy logic presents a promising frontier of overcoming intrinsic rigidities inscribed in ontologies like OWL while also conserving robust logical inference. As KGs become intractably complex, quantum computing can help solve scalability and computational intractability by leveraging states of superposition and entanglement. Challenges remain in devising quantum neural networks (QNN) on account of the fact that the quantum formalism is inherently linear whereas ANNs require nonlinear activation functions \cite{beer2020training, schuld2014quest}. Pending advances in QNNs, analogous uses of quantum computation in the field of biological simulation have been proposed and developed \cite{peruzzo2014variational, aspuru2005simulated}. The latter path represents the converse frontier of knowledge representation: physical system simulation. Jointly, neuro-symbolic representation and physical system simulation constitute a powerful amalgam within the domain of knowledge representation. 

\section{Conclusion}
A host of connected themes thread our survey of hybridizing ontologies and artificial neural networks. Ontologies represent an extension of Symbolic AI, also termed Good-Old-Fashioned AI (GOFAI), which models artificial intelligence on symbolic reasoning. The rival method of artificial intelligence includes the connectionist paradigm upon which artificial neural networks (ANNs) rest. Whereas GOFAI models reasoning as a top-down process of logical inference, ANNs model reasoning as a bottom-up process of pattern extraction or generalization. Separately these contrasting models entail the following functional trade-off: GOFAI can reason robustly but cannot adapt, whereas ANNs can learn and adapt from new data by updating their internal models but cannot reason robustly. Due to these cross-advantages, we have proposed applying neural dense-vector embeddings to ontologies in order to encode latent semantic structures that make them adaptable, interoperable and extensible. In addition, we have proposed techniques that preserve their logical structure in order to facilitate classification, prediction, scalability and search. 

Embeddings involve a novel technique that leverages ANNs to encode words or symbols as dense vectors in n-dimensional space in order to represent similarity relationships. The application of embeddings to ontological frameworks such as OWL and domain-specific ontologies such as GO show promising results in extending the application of ontologies to real-world problems by enhancing knowledge representation and retrieval. Graph-Neural Networks provide another avenue of achieving similar results through end-to-end learning. Despite revealing latent semantic relationships and enhancing contextual richness, these methods trade-off explicit semantics and inference capabilities. 

In order to maintain the contextual gains of embeddings, but also increase the representational and reasoning power of ontologies, I have proposed hybridizing fuzzy logic and quantum computation. Fuzzy logic encodes uncertainty in classification tasks through membership gradience, whereas quantum computing enables simultaneous operations on fuzzy sets in states of superposition. While neuro-fuzzy systems have been developed, computational complexity remains an outstanding problem. Formal neuro-fuzzy-quantum systems, pending engineering advances in quantum computing, would solve this and other outstanding problems with KGs such as interoperability and graph completion. Finally, the next frontier of ontologies and KGs will be in the realm of complexity modeling, where computational intractability becomes a problem. For this reason, formalizing fuzzy-quantum systems remains a promising avenue of solving outstanding problems in knowledge representation. 

In summation, the gains in semantics, symbol and statistical modeling, explainability, and relational structure of ontologies produce losses in feature engineering, scalability, and adaptability. Conversely, the scalability, generalization, flexibility, and automated feature learning of ANNs produce losses in interpretability, semantic loss/inference loss, and impose huge data requirements. To face the complexity problems of knowledge representation we need both frameworks.

\bibliography{bibliography}
\end{document}